\documentclass[conference]{IEEEtran}
\IEEEoverridecommandlockouts
\usepackage{cite}
\usepackage{amsmath,amssymb,amsfonts}
\usepackage{algorithmic}
\usepackage{graphicx}
\usepackage{textcomp}
\usepackage{xcolor}
\usepackage{enumitem} 
\usepackage{etoolbox}
\usepackage{booktabs}
\usepackage{float}   
\usepackage{makecell}
\usepackage{colortbl}
\usepackage{color}
\definecolor{mygray}{gray}{.9}

\makeatletter
\patchcmd{\@makecaption}
  {\scshape}
  {}
  {}
  {}
\makeatother

\def\BibTeX{{\rm B\kern-.05em{\sc i\kern-.025em b}\kern-.08em
    T\kern-.1667em\lower.7ex\hbox{E}\kern-.125emX}}
\begin{document}

\title{HiF-DTA: Hierarchical Feature Learning Network for Drug–Target Affinity Prediction\protect\\
\thanks{Minghui’s work is supported in part by the National Natural Science Foundation of China (Grant No. 62572206). Shengshan’s work is supported in part by the National Natural Science Foundation of China (Grant No.62372196). The work is supported by the HPC Platform of Huazhong University of Science and Technology.

Shengqing is the corresponding author.}
}

\author{\IEEEauthorblockN{1\textsuperscript{st} Minghui Li}
\IEEEauthorblockA{\textit{School of Software Engineering} \\
\textit{Huazhong University of}\\
\textit{Science and Technology}\\
Wuhan, China \\
minghuili@hust.edu.cn}
\protect\\
\IEEEauthorblockN{4\textsuperscript{th} Wei Wan}
\IEEEauthorblockA{
\textit{Faculty of Data Science} \\
\textit{City University of Macau}\\
Macau, China \\
weiwan@cityu.edu.mo}
\and
\IEEEauthorblockN{2\textsuperscript{nd} Yuanhang Wang}
\IEEEauthorblockA{\textit{School of Software Engineering} \\
\textit{Huazhong University of}\\
\textit{Science and Technology}\\
Wuhan, China \\
wangyuanhang@hust.edu.cn}
\protect\\
\IEEEauthorblockN{5\textsuperscript{th} Shengshan Hu}
\IEEEauthorblockA{
\textit{School of Cyber Science and Engineering} \\
\textit{Huazhong University of}\\
\textit{Science and Technology}\\
Wuhan, China \\
hushengshan@hust.edu.cn}
\and
\IEEEauthorblockN{3\textsuperscript{rd} Peijin Guo}
\IEEEauthorblockA{
\textit{School of Cyber Science and Engineering} \\
\textit{Huazhong University of}\\
\textit{Science and Technology}\\
Wuhan, China \\
gpj@hust.edu.cn}
\protect\\
\IEEEauthorblockN{6\textsuperscript{th} Shengqing Hu}
\IEEEauthorblockA{
\textit{Union Hospital,} \\
\textit{Tongji Medical College} \\
\textit{Huazhong University of}\\
\textit{Science and Technology}\\
Wuhan, China \\
hsqha@126.com}
}

\maketitle

\begin{abstract}
Accurate prediction of Drug-Target Affinity (DTA) is crucial for reducing experimental costs and accelerating early screening in computational drug discovery. While sequence-based deep learning methods avoid reliance on costly 3D structures, they still overlook simultaneous modeling of global sequence semantic features and local topological structural features within drugs and proteins, and represent drugs as flat sequences without atomic-level, substructural-level, and molecular-level multi-scale features. We propose HiF-DTA, a hierarchical network that adopts a dual-pathway strategy to extract both global sequence semantic and local topological features from drug and protein sequences, and models drugs multi-scale to learn atomic, substructural, and molecular representations fused via a multi-scale bilinear attention module. Experiments on Davis, KIBA, and Metz datasets show HiF-DTA outperforms state-of-the-art baselines, with ablations confirming the importance of global-local extraction and multi-scale fusion.
\end{abstract}


\section{Introduction}
Accurate prediction of drug–target affinity (DTA) is essential for drug screening, immune modulation and precision medicine. Experimental assays are costly and slow, driving the adoption of deep-learning models that operate in an end-to-end manner.
Existing deep-learning methods fall into two streams: structure- and sequence-based. The former hinges on 3-D coordinates that are scarce and expensive to obtain \cite{3Ddrug,3Dprotein,acquisition3D,guo2025multi}, whereas the latter relies solely on SMILES or amino-acid strings and is more scalable \cite{li2025mvsf,li2024vidta}. 

Nevertheless, current sequence-based models have three common drawbacks:
(1) \textit{Local topology neglect}: global semantics are emphasised, but binding-site or substructural cues are overlooked \cite{deepdta}.
(2) \textit{Insufficient multi-scale modelling}: atom- or residue-level features are used, yet intermediate substructures are rarely exploited \cite{meng2024fusiondti,li2025hcaf}.
(3) \textit{Modality isolation}: CNN or GNN encoders process sequences or graphs separately, leaving semantics and structures unpaired (e.g., DeepDTA \cite{deepdta}, GraphDTA \cite{nguyen2021graphdta}).

We present HiF-DTA, a hierarchical feature-learning network that addresses the above issues. A dual-pathway encoder captures global sequence semantics (BiLSTM for drugs, Mamba for proteins) and local topology (PNA-based GNN on molecular/residue graphs). For drugs, atomic, substructural and molecular features are explicitly extracted and fused by multi-scale bilinear attention, enabling fine-grained interaction modelling.

The main contributions of this work are as follows:
\begin{itemize}[leftmargin=*,label=\textbullet,topsep=0pt,partopsep=0pt,itemsep=0pt,parsep=0pt]
\item HiF-DTA integrates global and local cues via dual-pathway encoding for both drugs and proteins.
\item Three-level drug features (atom, substructure, molecule) are jointly learned and bilinearly attended to protein clusters.
\item State-of-the-art results on Davis, KIBA and Metz with ablations verifying the value of global-local and multi-scale designs.
\end{itemize}

\section{MATERIALS AND METHODOLOGY}

\begin{figure*}[t]
    \centering
    \includegraphics[width=\textwidth]{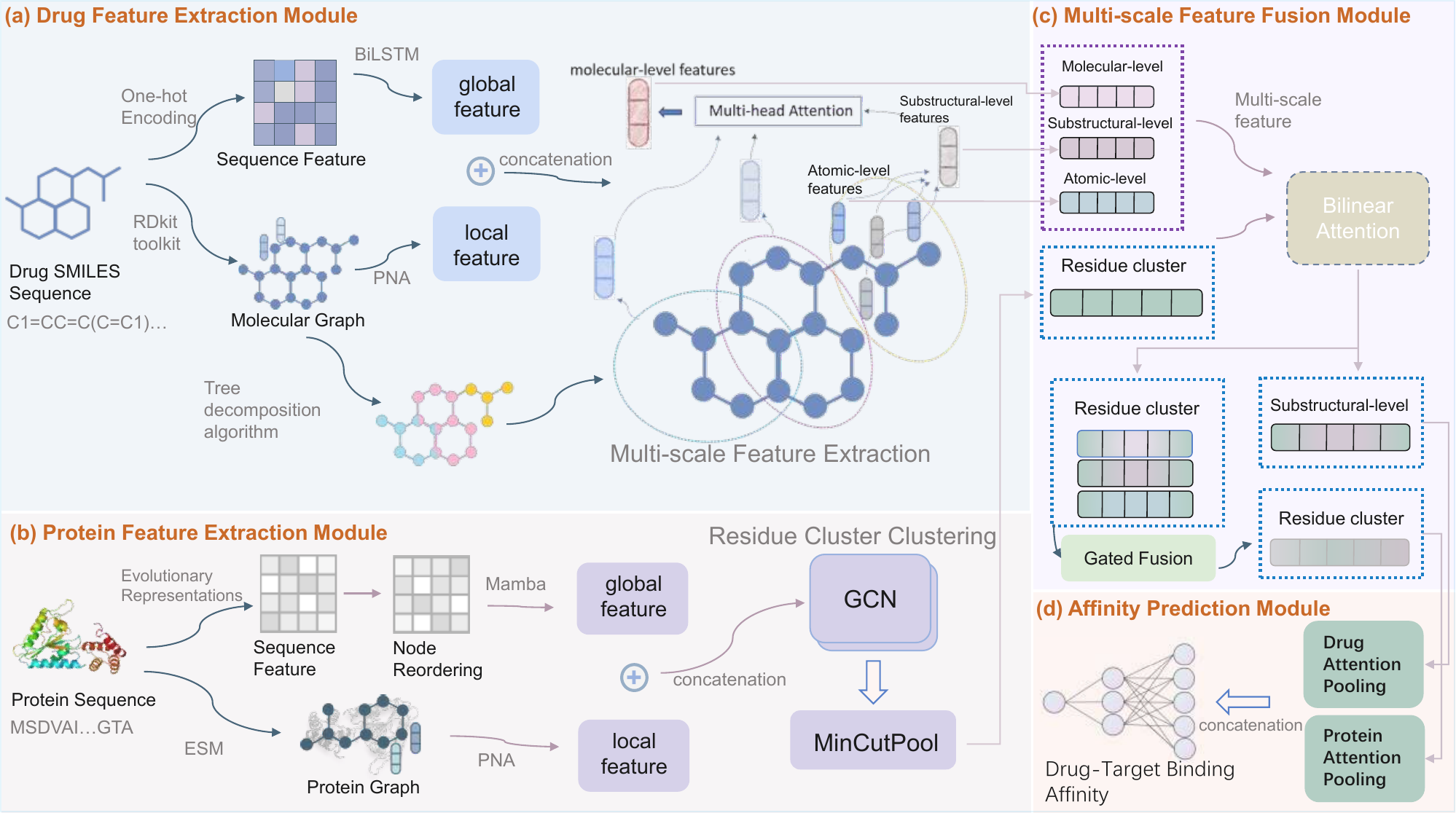}
    \caption{Pipeline of the proposed HiF-DTA scheme.}
    \label{fig:pipeline}
\vspace{-3mm}
\end{figure*}

\subsection{Datasets}
We evaluated the performance of our model on three widely used benchmark datasets: Davis \cite{davis2011comprehensive}, Metz \cite{metz2011navigating}, and KIBA \cite{tang2014making}. See Table~\ref{tab:1} for details. In these datasets, a smaller $K_d$ value indicates a stronger binding affinity between the drug and the target. To reduce the variance, the $K_d$ values in the Davis dataset are commonly transformed into a logarithmic scale using the following formula:
\begin{equation}
p K_{d}=-\lg \left(\frac{K_{d}}{10^{9}}\right)
\end{equation}

\subsection{Overview}
HiF-DTA comprises four modules: drug extractor, protein extractor, fusion unit, and affinity predictor (Figure~\ref{fig:pipeline}).

The drug extractor encodes SMILES into one-hot sequences and RDKit graphs, extracts global and local features via BiLSTM and PNA, applies tree decomposition for atomic, substructural, and molecular levels, and integrates them through multi-head attention into a multi-scale drug tensor.

The protein extractor derives ESM2 embeddings, enhances global semantics using Mamba, constructs a residue graph processed by PNA for local details, and refines the combined features via a two-layer GCN and MinCutPool to obtain residue-cluster representations.

The fusion unit interacts multi-scale drug and residue-cluster features through bilinear attention, updates substructural descriptors, and integrates multi-scale cluster representations via gated fusion into a unified embedding.

The affinity predictor performs attentive pooling on unified and substructural features, concatenates them, and predicts binding affinity via a fully connected network.

\begin{table}[t]
\centering
\caption{Statistical Analysis of Benchmark Datasets}
\setlength{\tabcolsep}{14.0pt}
\scalebox{0.99}{
\begin{tabular}{l|c|c|c}
\toprule[0.15em]
\rowcolor{mygray} Dataset & Drugs & Proteins & Affinities \\
\midrule[0.1em]
Davis & 68 & 442 & 30056 \\ 
Metz & 240 & 121 & 13669 \\ 
KIBA & 2111 & 229 & 118254 \\
\bottomrule[0.15em]
\end{tabular}}
\label{tab:1}
\vspace{-4mm}
\end{table}

\subsection{Drug Feature Extraction Module}
\textit{1) Sequence Feature and Molecular Graphs Representation:} The module takes SMILES strings as input. Each atom is one-hot encoded into a 43-dimensional vector capturing connectivity, hydrogen count, hybridization and aromaticity,constituting the sequence feature
$\mathbf{h}_{\text{sequence}}$ \cite{koh2023psichic,guo2025uncertainty}. We also parse the SMILES strings using the RDKit toolkit into a molecular graph $G = (V, E)$.

\textit{2) Global and Local Features Extraction:} For the global sequence feature, we map the sequence feature vector $\mathbf{h}_{\text {sequence}}$ through two fully connected layers with two activation functions and a final normalization to obtain the initial feature $\mathbf {\hat{h}}_{\text {Atom}}$ of the drug sequence.

\begin{equation}
\hat{\mathbf{h}}_{\text {Atom}}=\operatorname{Norm}\left(\operatorname{\sigma}\left(\operatorname{\sigma}\left(\mathbf{h}_{\text {sequence}} {W}_{1}+{b}_{1}\right) {W}_{2}+{b}_{2}\right)\right)
\end{equation}

where ${W}_1$ and ${W}_2$ are the weight matrices of the first and second fully connected layers, respectively, and $b_1$, $b_2$ are the corresponding bias vectors. The activation function $\sigma(\cdot)$ denotes the ReLU nonlinearity, and $\operatorname{Norm}(\cdot)$ represents layer normalization applied to the final output.
Then, we extract contextual feature $\mathbf{h}_{\text {BiLSTM}}$ of the global sequence using a BiLSTM \cite{siami2019performance}.
\begin{equation}
\mathbf{h}_{\text {BiLSTM}}=\operatorname{BiLSTM}\left(\mathbf {\hat{h}}_{\text {Atom}}\right)
\end{equation}

\begin{figure}[t]
    \begin{minipage}[b]{0.5\textwidth} 
        \centering
        \includegraphics[width=\linewidth]{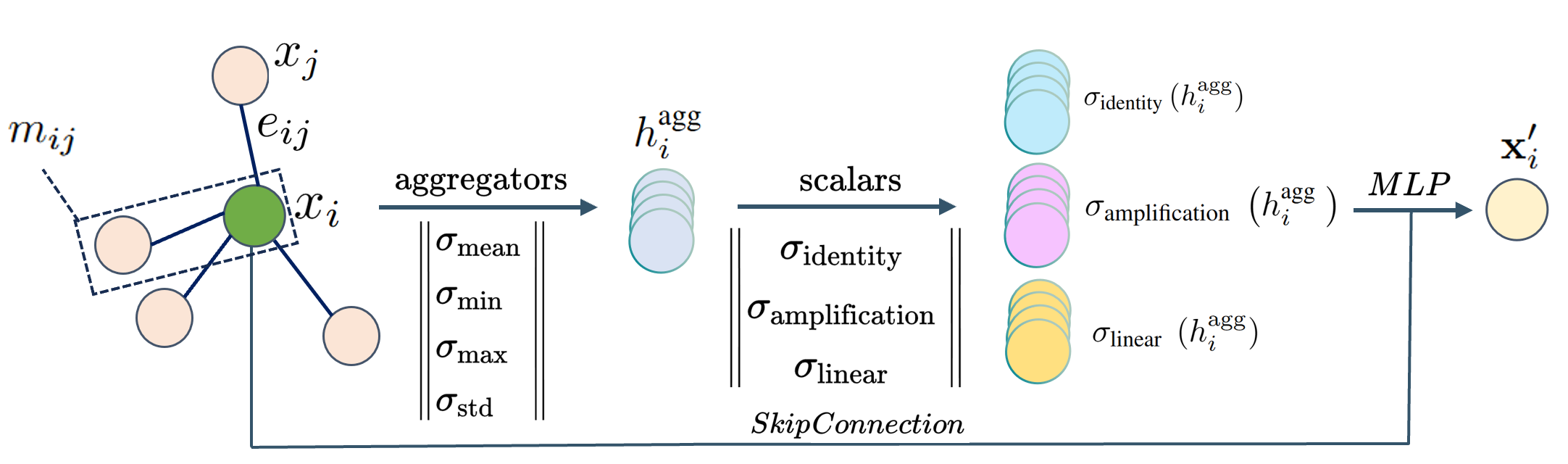} 
        \caption{Architecture of the Principal Neighbourhood Aggregation (PNA).}
\label{fig2}
\end{minipage}
\vspace{-4mm}
\end{figure}

For local structural modeling, we use the Principal Neighborhood Aggregation (PNA) \cite{corso2020principal} within a Message Passing Neural Network (MPNN) to capture atom interactions in molecular graphs (Fig. \ref{fig2}). For atom $i$ and neighbor $j$ with features $x_i$, $x_j$, and bond feature $e_{ij}$, the message $m_{ij}$ is computed by concatenating these features and projecting them into hidden dimension $d$ through $\operatorname{MLP}_{\mathrm{pre}}$.
\begin{equation}
m_{i j}=\operatorname{MLP}_{\mathrm{pre}}\left(\left[x_{i}\left\|x_{j}\right\| \phi\left(e_{i j}\right)\right]\right)
\end{equation}
where $\|$ denotes feature concatenation operation, and $\phi\left(\cdot\right)$ is an edge encoder mapping edge features to the hidden dimension. The local feature of node $i$ aggregates messages $m_{ij}$ from neighbors $j$, followed by scaling with aggregators $A=\{\sigma_{\text{mean}}, \sigma_{\text{min}}, \sigma_{\text{max}}, \sigma_{\text{std}}\}$ and scalers $S=\{\sigma_{\text{identity}}, \sigma_{\text{amplification}}, \sigma_{\text{linear}}\}$, as:
\begin{equation}
{h}_{i}=\bigoplus_{s \in S} s\left(\bigoplus_{a \in A} a\left(\left\{{m}_{i j} \mid j \in \mathcal{N}(i)\right\}\right)\right)
\end{equation}
where $\bigoplus$ is the concatenation operation, and $\mathcal{N}(i) $ denotes the aggregation of the set of neighbor nodes of node $i$.

The original features ${x}_{i}$ are concatenated with the local features ${h}_{i}$, followed by a nonlinear transformation through the $\mathrm{MLP}_{\text {post}}$, which reduces the dimensionality to $d$. A subsequent linear projection is then applied to generate the final node representation $\mathbf{x}_{i}^{\prime}$:
\begin{equation}
\mathbf{x}_{i}^{\prime}=\mathrm{W} \cdot \mathrm{MLP}_{\text {post}}\left(\left[\mathbf{x}_{i} \| {h}_{i}\right]\right)
\end{equation}

Through message passing, each node representation is updated to yield the local atom feature $\mathbf{h}{\text{MPNN}}$. This feature is concatenated with the global atom feature $\mathbf{h}{\text{BiLSTM}}$ and refined via linear mapping, activation, and layer normalization to form the final atom feature $\mathbf{h}_{\text{Atom}}$.
\begin{equation}
\mathbf{h}_{\text {Atom}}=\operatorname{Norm}\left(\sigma\left(\mathrm{W}\left[\mathbf{h}_{\text{MPNN}} \| \mathbf{h}_{\text {BiLSTM}}\right]+b\right)\right)
\end{equation}
where, $\operatorname{Norm}(\cdot)$ represents layer normalization applied to the final output, $\sigma(\cdot)$ denotes the ReLU activation function, $\mathrm{W}$ is the weight matrix that projects the concatenated features from dimension $2d$ down to $d$, and $b$ represents the bias vector.

\begin{figure}[t]
    \begin{minipage}[b]{0.5\textwidth} 
        \centering
        \includegraphics[width=\linewidth]{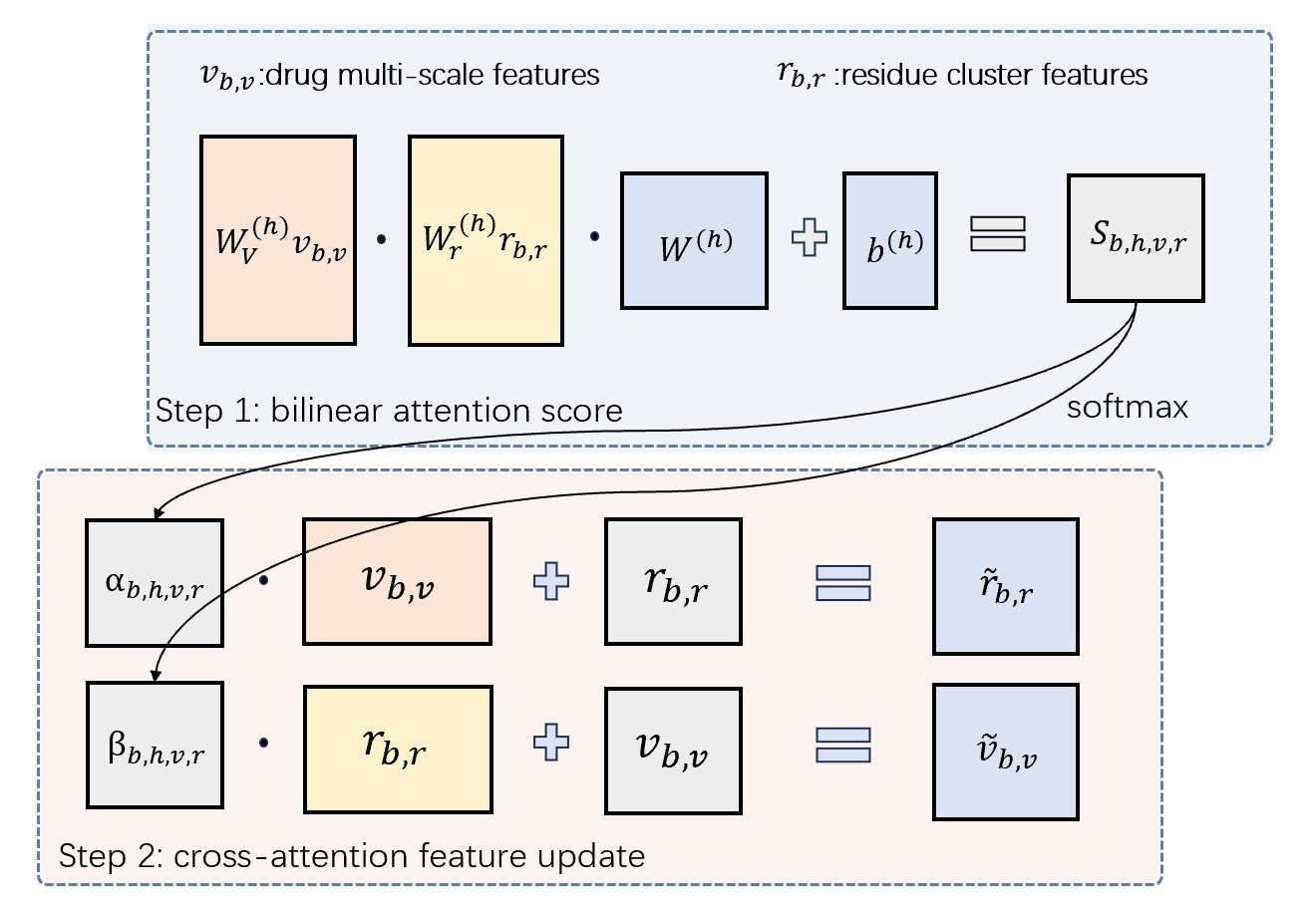} 
        \caption{Illustration of the Multi-scale Feature Fusion Module.}
\label{fig3}
\end{minipage}
\vspace{-3mm}
\end{figure}

\textit{3) Multi-scale Feature Extraction:}
The tree decomposition algorithm \cite{jin2018junction} partitions the molecular graph into substructures, then based on the atom-to-substructure mapping $\operatorname{map}_{v \rightarrow c}$, the substructural features are updated by performing mean pooling over the atomic features $\mathrm{\tilde{H}_{c}}$:
\begin{equation}
\mathrm{\tilde{H}_{c}}=\operatorname{MeanPool}\left(\mathbf{h}_{\text {Atom}}, \operatorname{map}_{v \rightarrow c}\right) \in R^{C \times d}
\end{equation}
MeanPool averages features over C substructures. Type-based label embeddings first yield initial substructural features $\mathbf{h}_{\text {Group}}$; $\mathrm{\tilde{H}_{c}}$ then passes through a linear layer plus activation and joins $\mathbf{h}_{\text {Group}}$ via a residual link to produce updated substructural features $\mathrm{H_{c}}$.
\begin{equation}
 \mathrm{H_{c}}=\mathbf{h}_{\text {Group}}+\operatorname{ReLU}\left(W \cdot \mathrm{\tilde{H}_{c}}\right) \in R^{C \times d}
\end{equation}

$ \mathrm{H_{c}}$ is reshaped into $ h$ attention heads of dimension $ d/h$, and each head computes an attention score $\boldsymbol{\alpha}_{i}$ via an independent multi-layer perceptron:
\begin{equation}
\boldsymbol{\alpha}_{i}=\operatorname{MLP}_{i}\left(\mathrm{{H}_{c}^{(i)}}\right) \in \mathbb{R}^{C \times 1}, \quad i=1, \ldots, h
\end{equation}

The attention scores are regularized with dropout and normalized per-molecule via batch-wise $\operatorname{batch_c}$ and Softmax to obtain $\hat{\boldsymbol{\alpha}}_{i}$:
\begin{equation}
\hat{\boldsymbol{\alpha}}_{i}=\operatorname{Softmax}\left(\operatorname{Dropout}\left(\boldsymbol{\alpha}_{i}\right), \operatorname{batch_c}\right)
\end{equation}

The attention weights from all heads are concatenated to form a unified attention vector $\hat{\boldsymbol{\alpha}}$:
\begin{equation}
\hat{\boldsymbol{\alpha}}=\left[\hat{\boldsymbol{\alpha}}_{1}\left\|\hat{\boldsymbol{\alpha}}_{2}\right\| \cdots \| \hat{\boldsymbol{\alpha}}_{h}\right]
\end{equation}

This attention vector is applied to the substructural features via element-wise multiplication. A batch-wise weighted sum pooling is then performed to generate the fused molecular representation:
\begin{equation}
\mathbf{h}_{\mathrm{mol}}=\operatorname{ScatterSum}\left(\hat{\boldsymbol{\alpha}} \odot \mathrm{H_{c}}, \operatorname{batch_c}\right)
\end{equation}
where $\odot$ denotes element-wise multiplication and $\operatorname{ScatterSum}$ aggregates the weighted features within each batch by summing them according to the batch indices.

{\begingroup         
\renewcommand{\floatpagefraction}{.9}   
\makeatletter
\@beginparpenalty=10000     
\makeatother

\begin{table}[H]
\centering  
\caption{Hyperparameter Settings}  
\label{tab:2}
\setlength{\tabcolsep}{8.0pt}
\scalebox{0.99}{
\begin{tabular}{l | l}
\toprule[0.15em]
\rowcolor{mygray} Hyper-parameters & Setting \\ 
\midrule[0.15em]
Learning rate & \{1e-3, 5e-4\}   \\ 
Batch size & 64 \\ 
Epoch & 400 \\ 
Optimizer & Adam \\ 
total\_layer & 3 \\ 
hidden\_channels & 200 \\ 
number of residue clusters per layer & [5, 10, 20] \\
Heads number of the Bilinear cross-attention & 4 \\ 
Heads of Drug Multi-scale feature updating   & 4 \\ 
Dropout of Drug Multi-scale feature updating & 0.2 \\ 
\bottomrule[0.1em]
\end{tabular}}
\end{table}

\vspace{-5mm}

\begin{table}[H]
\centering  
\caption{Comparison Results on the Davis Benchmark Dataset}  
\label{tab:3}
\begin{tabular}{l|l|l|l|l}
\toprule[0.15em]
\rowcolor{mygray} Method            & \textbf{$\mathbf{\mathrm{CI} \uparrow}$}             & $r_{m}^{2}\uparrow$            & \textbf{$\mathrm{PCC} \uparrow$}           & \textbf{$\mathrm{MSE} \downarrow$}           \\ 
\midrule[0.15em]
GraphDTA          & 0.888          & 0.699          & 0.8475         & 0.232          \\ 
AttentionDTA      & 0.8947         & 0.7404         & 0.8721         & 0.1912         \\ 
ColdDTA           & 0.8938         & 0.7606         & 0.8861         & 0.1695         \\ 
AttentionMGT-DTA  & 0.891          & 0.699          &  -              & 0.193          \\ 
GOaidDTA          & 0.891          & 0.654          & 0.85           & 0.229          \\ 
TF-DTA            & 0.8856         & 0.6703         &  -              & 0.2312         \\ 
TEFDTA            & 0.8925         & 0.7403         & 0.8617         & 0.21           \\ 
DGDTA             & 0.899          & 0.702          &  -              & 0.225          \\ 
PSICHIC           & 0.884          & 0.7262         & 0.8802         & 0.1783         \\ \midrule
HiF-DTA & \textbf{0.9026} & \textbf{0.762} & \textbf{0.8921} & \textbf{0.1654} \\ \bottomrule
\end{tabular}
\vspace{-3mm}
\end{table}

\endgroup}

\subsection{Protein Feature Extraction Module}
\textit{1) Sequence Feature and protein Graphs Representation:} ESM2 33rd-layer embeddings of size $R \times 1280$ are taken as evolutionary features. Thresholded ESM2 contact probabilities define the protein graphs $G=(R, L)$ with residues as nodes and contacts as edges. Sequence features concatenate these embeddings with a residue matrix that combines 21-dimensional one-hot encoding and twelve physicochemical descriptors. The contact matrix is converted to edge features through radial basis function mapping.

\textit{2) Global and Local Features Extraction:}
Each protein node is associated with a batch index denoted as $\operatorname{batch_R}$. First, the nodes are reordered based on their node degree $d_i$ and batch index $\operatorname {batch_R}$, and the resulting permutation is recorded as $\pi$:
\begin{equation}
\pi=\operatorname{argsort}\left(\operatorname{array}\left[d_{i}, \operatorname{batch_R}\right]\right)
\end{equation}

The reordered node features are obtained, and the sparse node features are converted into dense matrices $\mathrm{H}$ according to batches:
\begin{equation}
\mathrm{H} \in \mathbb{R}^{B \times V_{m} \times d}
\end{equation}

where $V_m$ is the maximum number of nodes within a batch, and missing positions are filled with masks. Based on the Mamba structured state space model, the global interaction features $\mathrm{H}^{\prime}$ of nodes are computed as:
\begin{equation}
\mathrm{H}^{\prime}=\mathrm{Mamba}(\mathrm{H})
\end{equation}

Then, the sparse valid nodes are extracted, and the inverse permutation is applied to restore the original node order:
\begin{equation}
\mathbf{R}_{\text {Mamba}}=\mathrm{H}^{\prime}[\mathrm{mask}]_{\pi^{-1}}
\end{equation}

{\begingroup          
\renewcommand{\floatpagefraction}{.9}   
\makeatletter
\@beginparpenalty=10000     
\makeatother

\begin{table}[H]
\centering  
\caption{Comparison Results on the Metz Benchmark Dataset} 
\label{tab:4}
\begin{tabular}{l|l|l|l|l}
\toprule[0.15em]
\rowcolor{mygray} Method           & \textbf{$\mathbf{\mathrm{CI} \uparrow}$}             & $r_{m}^{2}\uparrow$            & \textbf{$\mathrm{PCC} \uparrow$}           & \textbf{$\mathrm{MSE} \downarrow$}           \\ 
\midrule[0.15em]
GraphDTA          & 0.8621         & 0.7079         & 0.8548         & 0.1714         \\ 
ArKDTA            & 0.843          &   -             &  -              & 0.1703         \\ 
AttentionDTA      & 0.8755         & 0.7048         & 0.8565         & 0.1612         \\ 
ColdDTA           & 0.8738         & 0.7116         & 0.8622         & 0.1553         \\ 
TEFDTA            & 0.8445         & 0.6171         & 0.8376         & 0.1873         \\ 
PSICHIC           & 0.8751         & 0.7234         & 0.8688         & 0.1469         \\  \midrule
HiF-DTA & \textbf{0.8831} & \textbf{0.7486} & \textbf{0.8774} & \textbf{0.1369} \\ \bottomrule
\end{tabular}
\end{table}

\vspace{-5mm}

\begin{table}[H]
\centering  
\caption{Comparison Results on the KIBA Benchmark Dataset}  
\label{tab:5}
\begin{tabular}{l|l|l|l|l}
\toprule[0.15em]
\rowcolor{mygray} Method            & \textbf{$\mathbf{\mathrm{CI} \uparrow}$}             & $r_{m}^{2}\uparrow$            & \textbf{$\mathrm{PCC} \uparrow$}           & \textbf{$\mathrm{MSE} \downarrow$}           \\ 
\midrule[0.15em]
GOaidDTA                  & 0.876          & 0.706          & 0.868          & 0.179          \\ 
AttentionDTA              & 0.8799         & 0.735          & 0.8739         & 0.1668         \\ 
ColdDTA                   & 0.8689         & 0.7054         & 0.8671         & 0.1762         \\ 
TF-DTA                    & 0.8768         & 0.7344         &   -             & 0.1771         \\ 
TEFDTA                    & 0.8675         & 0.7065         & 0.8546         & 0.1864         \\ 
AttentionMGT-DTA          &  0.893           & 0.786        &  -              &  0.14         \\ 
PSICHIC & 0.8816 & 0.7571 & 0.8783 & 0.1556 \\ 
\midrule
HiF-DTA & \textbf{0.8948} & \textbf{0.7869} & \textbf{0.8947} & \textbf{0.1387} \\ \bottomrule
\end{tabular}
\vspace{-3mm}
\end{table}

\endgroup}

For local structure we reuse the atomic-graph MPNN to yield residue features, concatenate them with Mamba outputs, and fuse the pair through linear projection, ReLU and layer normalization to obtain the unified residue feature:
\begin{equation}
\mathbf{R}_{\text {residue}}=\text {Norm}\left(\sigma\left(W\left[\mathbf{R}_{\text {MPNN}} \| \mathbf{R}_{\text {Mamba}}\right]+b\right)\right)
\end{equation}
where $\operatorname{Norm}(\cdot)$ represents layer normalization applied to the final output, $\sigma(\cdot)$ denotes the ReLU activation function, $\mathrm{W}$ is the weight matrix that projects the concatenated features from dimension $2d$ down to $d$, and $b$ represents the bias vector.

\textit{3) Residue Cluster Clustering:} 
A two-layer GCN refines $\mathbf{R}_{\text{residue}}$, softmax with batch padding outputs assignment matrix $M \in \mathbb{R}^{B \times R_m \times cl}$, and dense mincut pooling algorithm pools \cite{bianchi2020spectral} $\mathbf{R}_{\text {residue}}$ via $M$ to yield the residue cluster feature matrix $\mathbf{R}_{\text {cluster}}\in \mathbb{R}^{B \times cl \times d}$.

\begin{table*}[t]
\centering  
\caption{Ablation Results on Different Multi-scale Features and Different Feature Fusion Strategies} 
\vspace{-2mm}
\label{tab:8}
\begin{tabular}{ccc|ccc|cccc}
\toprule[0.15em]
\rowcolor{mygray}
\multicolumn{3}{c|}{Multi-scale Features} & \multicolumn{3}{c|}{Fusion Strategy} & \multicolumn{4}{c}{Evaluation Metrics} \\ 
\midrule[0.15em]
Molecular & substructural & Atomic & Bilinear Attention & Concatenation   & Addition & \textbf{$\mathrm{CI} \uparrow$}             & $r_{m}^{2}\uparrow$            & \textbf{$\mathrm{PCC} \uparrow$}           & \textbf{$\mathrm{MSE} \downarrow$}             \\
$\checkmark$   &$\checkmark$      &$\checkmark$        & $\times$       & $\checkmark$     & $\times$      & 0.8733     &0.7205      &0.8551    &0.1691      \\
$\checkmark$   &$\checkmark$      &$\checkmark$        & $\times$       & $\times$     & $\checkmark$      &0.8544      &0.6867      &0.8371    & 0.1753     \\
$\checkmark$ &$\times$       &$\times$      &$\checkmark$       &$\times$         &$\times$      & 0.8754    & 0.7363   & 0.871   & 0.1483         \\
$\times$ &$\checkmark$       &$\times$      &$\checkmark$       &$\times$         &$\times$ &0.8793      &0.738      &0.8721    & 0.1433        \\
$\times$&  $\times$     & $\checkmark$     &$\checkmark$        & $\times$       & $\times$     &0.869       &0.7374      &0.8591      & 0.1481         \\     \midrule
$\checkmark$&$\checkmark$     &$\checkmark$        & $\checkmark$       & $\times$       &$\times$      & \textbf{0.8831} & \textbf{0.7486} & \textbf{0.8774} & \textbf{0.1369} \\ 
\bottomrule
\end{tabular}
\vspace{-3mm}
\end{table*}

\begin{table}[t]
\centering  
\caption{Ablation Results on Different Drug Feature Representation Network} 
\vspace{-2mm}
\label{tab:6}
\begin{tabular}{l|l|l|l|l}
\toprule[0.15em]
\rowcolor{mygray}
Drug Representation & \textbf{$\mathrm{CI} \uparrow$} & $r_{m}^{2}\uparrow$ & \textbf{$\mathrm{PCC} \uparrow$} & \textbf{$\mathrm{MSE} \downarrow$} \\ \midrule[0.15em]
Global Features Only & 0.8545 & 0.6867 & 0.837 & 0.175 \\
Local Features Only & 0.878 & 0.725 & 0.866 & 0.149 \\ \midrule
Both Global \& Local Features  & \textbf{0.8831} & \textbf{0.7486} & \textbf{0.8774} & \textbf{0.1369} \\ \bottomrule
\end{tabular}
\end{table}

\begin{table}[t]
\centering  
\caption{Ablation Results on Different Protein Feature Representation Network} 
\vspace{-2mm}
\label{tab:7}
\begin{tabular}{l|l|l|l|l}
\toprule[0.15em]
\rowcolor{mygray}
Protein Representation & \textbf{$\mathrm{CI} \uparrow$} & $r_{m}^{2}\uparrow$ & \textbf{$\mathrm{PCC} \uparrow$} & \textbf{$\mathrm{MSE} \downarrow$} \\ 
\midrule[0.15em]
Global Features Only & 0.8754 & 0.6854 & 0.8593 & 0.1563 \\
Local Features Only & 0.8809 & 0.7447 & 0.8681 & 0.1492 \\ \midrule
Both Global \& Local Features & \textbf{0.8831} & \textbf{0.7486} & \textbf{0.8774} & \textbf{0.1369} \\ \bottomrule
\end{tabular}
\vspace{-3mm}
\end{table}

\subsection{Multi-scale Feature Fusion Module}
Thereafter, residue cluster features $\mathbf{R}_{\text {cluster}}$ and drug multi-scale features enter a multi-head bilinear cross-attention fusion network shown in Fig.~\ref{fig3}.

For each attention head $h$, residue cluster features $r_{b,r}$ and drug multi-scale features $v_{b,v}$ from the $b$-th sample are first projected into a shared latent space using learnable projection matrices $W_{r}^{(h)}$ and $W_{v}^{(h)}$, respectively.
\begin{equation}
\hat{r}_{b, r}=W_{r}^{(h)}  r_{b, r}   \quad \hat{v}_{b, v}=W_{v}^{(h)}  v_{b, v}
\end{equation}
For each attention head $h$, the projected drug multi-scale features $\hat{v}_{b,v}$ and residue cluster features $\hat{r}_{b,r}$ are multiplied element-wise. These are further weighted by a learnable per-channel scalar vector $W^{(h)} \in \mathbb{R}^{d \cdot k}$ to obtain the bilinear attention score, $k$ denotes the compression factor of the low-rank pooling:
\begin{equation}
S_{b, h, v, r}=\hat{v}_{b, v} \cdot \hat{r}_{b, r} \cdot W_{}^{(h)}+b^{(h)}
\end{equation}
where $b^{(h)}$ is a learnable bias scalar.The attention weights $\alpha_{b, h, v, r}$ are then obtained by applying a softmax function across all residue clusters. 
\begin{equation}
\alpha_{b, h, v, r}=\operatorname{softmax}_{r}\left(S_{b, h, v, r}\right)=\frac{\exp \left(S_{b, h, v, r}\right)}{\sum_{r^{\prime}} \exp \left(S_{b, h, v, r^{\prime}}\right)} 
\end{equation}
\begin{equation}
\tilde{r}_{b, r}={r}_{b, r}+\sum_{v} \alpha_{b, h, v, r} \cdot {v}_{b, v}
\end{equation}
The attention weights $\beta_{b, h, v, r}$ are then obtained by applying a softmax function across all drug multi-scale features. Using these attention weights, the multi-scale feature of each drug is updated by aggregating information from all residue clusters, followed by a residual connection to retain its original features:
\begin{equation}
\beta_{b, h, v, r}=\operatorname{softmax}_{v}\left(S_{b, h, v, r}\right)=\frac{\exp \left(S_{b, h, v, r}\right)}{\sum_{v^{\prime}} \exp \left(S_{b, h, v^{\prime}, r}\right)}
\end{equation}
\begin{equation}
\tilde{v}_{b, v}={v}_{b, v}+\sum_{r} \beta_{b, h, v, r} \cdot {r}_{b, r}
\end{equation}

The interaction is repeated independently at the atomic, substructural and molecular levels of the drug to produce three updated residue-cluster features $r_{b, r}^{(\text{atom})}$, $r_{b, r}^{(\text{substructure})}$ and $r_{b, r}^{(\text{drug})}$, which are then fused into the final residue-cluster feature $\widetilde{R}_{b, r}$ through softmax-weighted averaging.
\begin{equation}
\widetilde{R}_{b, r}=a_{1} \widetilde r_{b, r}^{(\text {atom})}+a_{2} \widetilde r_{b, r}^{(\text {substructure})}+a_{3} \widetilde r_{b, r}^{(\text {drug})}
\end{equation}

\subsection{Affinity Prediction Module}
The affinity prediction module integrates multi-scale protein and drug features using attention-based pooling and an MLP.

Protein attention pooling computes residue-level attention via residue-cluster weights and the assignment matrix, aggregates residue features into a global representation, and refines it through a two-layer MLP.

For drugs, multi-scale attention across molecular, substructural, and atomic levels generates atomic attention scores, which are aggregated and transformed by a two-layer MLP into the final drug vector.

The resulting protein and drug vectors are concatenated and passed to an MLP to predict the binding affinity score.

\section{EXPERIMENT AND RESULT}
\subsection{Experiment Setting}
To ensure reliable results, five-fold cross-validation was applied. The learning rate started at 0.001 and decayed to 0.0005 after 100 epochs. All datasets used a batch size of 64 and were trained for up to 400 epochs, with early stopping if no loss reduction occurred within 100 epochs. Hyperparameter details are summarized in Table \ref{tab:2}.

\subsection{Evaluation Metrics}
Since DTA prediction is a regression task, we adopt the Concordance Index (CI), Modified Correlation Coefficient ($r_{m}^{2}$), Pearson Correlation Coefficient (PCC), and Mean Squared Error (MSE) as evaluation metrics. These metrics quantify both ranking consistency and absolute deviation.

\subsection{Experimental Results}
We compared our method with GraphDTA\cite{nguyen2021graphdta}, AttentionDTA\cite{zhao2022attentiondta}, ColdDTA\cite{fang2023colddta}, AttentionMGT-DTA\cite{wu2024attentionmgt}, GOaidDTA\cite{GOaidDTA}, TF-DTA\cite{tfdta}, TEFDTA\cite{li2024tefdta}, DGDTA \cite{zhai2023dgdta}, PSICHIC\cite{koh2023psichic}, and ArKDTA\cite{gim2023arkdta}.

As shown in Tables \ref{tab:3}, \ref{tab:4}, and \ref{tab:5}, HiF-DTA achieves the best performance across all three benchmark datasets. It consistently ranks highest across all four evaluation metrics: CI, $r_{m}^{2}$, PCC, and MSE.

On Davis, HiF-DTA achieves a CI of 0.9026—the first reported result to surpass the 0.9 threshold. On Metz, it lowers the MSE to 0.1369, outperforming the previous best benchmark by 1\%. On the large-scale and highly heterogeneous KIBA benchmark, HiF-DTA further establishes state-of-the-art performance with a PCC of 0.8947.

\subsection{Ablation Experiments}
We assessed the effect of different drug feature extraction networks on DTA prediction using the Metz dataset under three settings: global only, local only, and combined global–local features. Results in Table~\ref{tab:6} show that integrating both feature types yields superior performance. Similarly, Table~\ref{tab:7} compares protein extraction architectures under the same settings, confirming that joint modeling of global semantic and local structural features consistently improves prediction accuracy.

We further examined the contribution of multi-scale features—atomic, substructural, and molecular levels—and compared fusion strategies including bilinear attention, concatenation, and addition. As shown in Table~\ref{tab:8}, our multi-scale fusion module achieves the best results across all metrics, with the substructural level performing best among single-scale variants.

\section{CONCLUSION}
This paper proposed HiF-DTA, a hierarchical feature learning network for DTA prediction. By adopting a dual-pathway encoding strategy, HiF-DTA integrates global sequence semantics and local structural features of drugs and proteins. It further captures multi-scale drug–target interactions (atomic, substructural, and molecular levels) and fuses them via a bilinear attention mechanism to enhance interaction representation. Experiments on three benchmark datasets show that HiF-DTA consistently surpasses state-of-the-art baselines, highlighting its potential in computational drug discovery.

\bibliographystyle{unsrt}
\bibliography{reference.bib}

\end{document}